\RequirePackage[bookmarksnumbered,unicode]{hyperref}
\documentclass[sigconf]{acmart}

\AtBeginDocument{%
  } 

\setcopyright{acmcopyright}
\copyrightyear{2018}
\acmYear{2018}
\acmDOI{XXXXXXX.XXXXXXX}
  
\acmPrice{15.00}
\acmISBN{978-1-4503-XXXX-X/18/06}
\renewcommand\footnotetextcopyrightpermission[1]{} 
\usepackage{enumitem}
\usepackage{amsmath,amsfonts}
\usepackage{amsthm}

\usepackage{hyperref}
\usepackage{xcolor}
\usepackage{footnote}
\usepackage{multirow}  
\usepackage{algorithm}
\usepackage{algpseudocode}
\usepackage{booktabs}
\usepackage{colortbl}
\usepackage{multirow}
\usepackage{footnote}
\usepackage{tablefootnote}
\usepackage{float}
\usepackage{threeparttable}
\usepackage{pifont}
\usepackage{textcomp}
\usepackage{tabularx}
\usepackage{graphicx}

\usepackage{listings}

\hypersetup{
  colorlinks,
  citecolor=violet,
  linkcolor=red,
  urlcolor=blue}

\begin{document}

\title{KEEP: A \underline{K}V-Cache-Centric Memory Management System for \underline{E}fficient \underline{E}mbodied \underline{P}lanning }

\settopmatter{printacmref=false,  printccs=false,  printfolios=false}
\begin{abstract}

Memory-augmented Large Language Models (LLMs) have demonstrated remarkable capability for complex and long-horizon embodied planning. By keeping track of past experiences and environmental states, memory enables LLMs to maintain a global view, thereby avoiding repetitive exploration. However, existing approaches often store the memory as raw text, leading to excessively long prompts and high prefill latency. While it is possible to store and reuse the KV caches, the efficiency benefits are greatly undermined due to frequent KV cache updates.
In this paper, we propose KEEP, a KV-cache-centric memory management system for efficient embodied planning. KEEP features 3 key innovations: (1) \textit{a Static-Dynamic Memory Construction} algorithm that reduces KV cache recomputation by mixed-granularity memory group; (2) a \textit{Multi-hop Memory Re-computation} algorithm that dynamically identifies important cross-attention among different memory groups and reconstructs memory interactions iteratively; (3) a \textit{Layer-balanced Memory Loading} that eliminates unbalanced KV cache loading and cross-attention computation across different layers. 
Extensive experimental results have demonstrated that KEEP achieves 2.68$\times$
speedup with negligible accuracy loss compared with text-based memory methods on ALFRED dataset. 
Compared with the KV re-computation method CacheBlend (EuroSys'25), KEEP shows 4.13\% success rate improvement and 1.90$\times$ time-to-first-token (TTFT) reduction.
Our code is available on \href{https://github.com/PKU-SEC-Lab/KEEP_Embodied_Memory}{https://github.com/PKU-SEC-Lab/KEEP\_Embodied\_Memory}.

\end{abstract}


\author{Zebin Yang$^{1,2}$, Tong Xie$^{1,2}$, Baotong Lu$^{4}$, Shaoshan Liu$^{3}$, Bo Yu$^{3}$\footnotemark[1], Meng Li$^{1,2,5}$\footnotemark[1]}

\affiliation{%
  \institution{$^1$Institute for Artificial Intelligence, $^2$School of Integrated Circuits, Peking University, Beijing, China}
  \country{}
}
\affiliation{%
  \institution{$^3$Shenzhen Institute of Artificial Intelligence and Robotics for Society, Shenzhen, China}
  \country{}
}
\affiliation{%
  \institution{$^4$Microsoft Research, $^5$Beijing Advanced Innovation Center for Integrated Circuits, Beijing, China}
  \country{}
}


\maketitle
\renewcommand{\thefootnote}{\fnsymbol{footnote}}
\footnotetext[1]{Corresponding author. Emails: boyu@cuhk.edu.cn, meng.li@pku.edu.cn}
\renewcommand{\thefootnote}{\arabic{footnote}}
\setcounter{footnote}{0}

\section{Introduction}
\label{sec:intro}

\begin{figure}[!tb]
    \centering
    \includegraphics[width=1.0\linewidth]{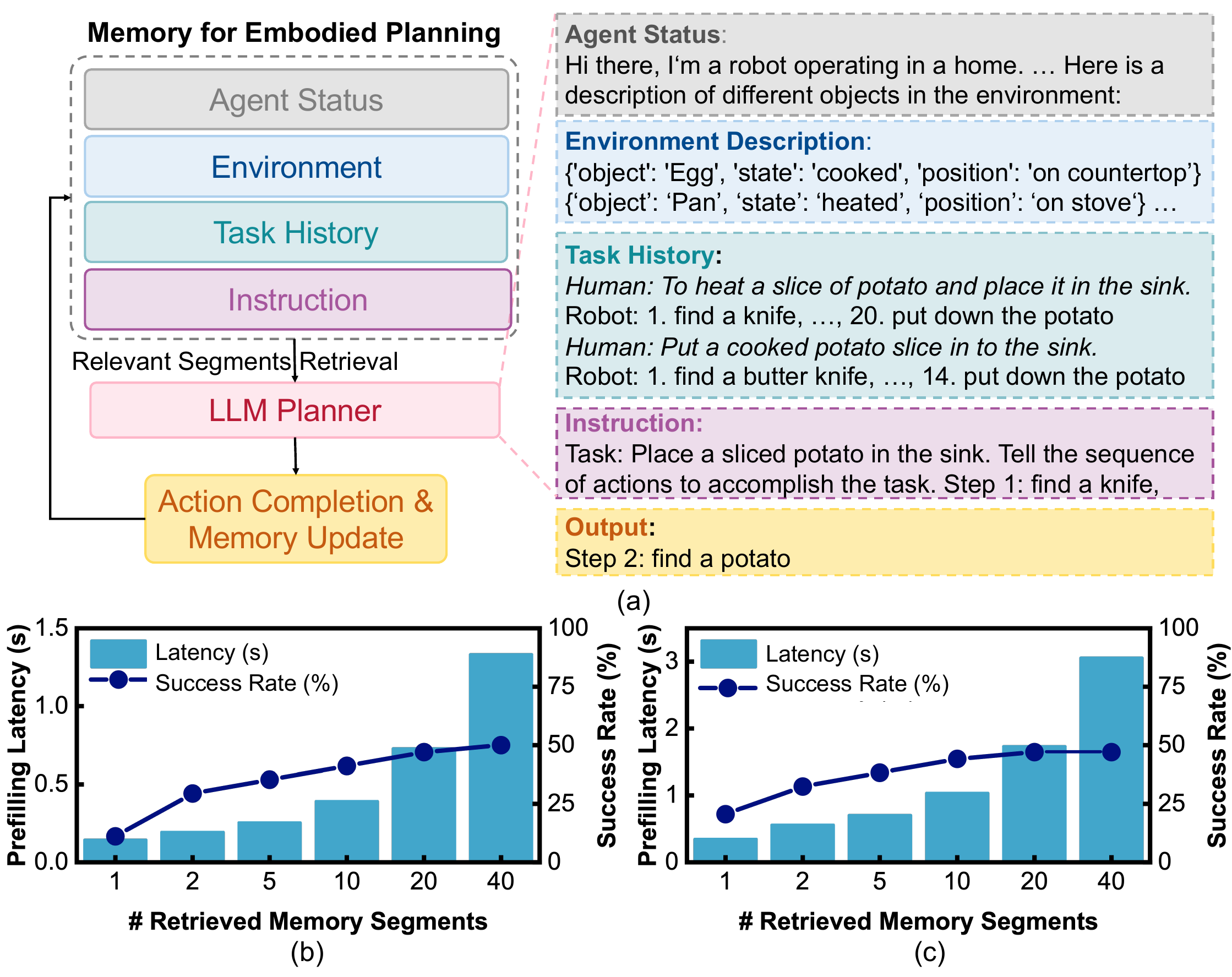}
    \caption{(a) An example of embodied planning with LLM planner. 
    In each step, a prompt composed of retrieved memory and instruction is given to the LLM planner.
     With the number of retrieved memory segments increasing, the success rate and prefilling latency both increase, evaluated on ALFRED dataset with (b) Qwen-14B and (c) Qwen-32B (INT4).
    }
    \label{fig:intro:llm_as_planner}
    \vspace{-7pt}
\end{figure}

Embodied planning tasks agents predicting sequences of actions to achieve long-term goals in real-world scenarios \cite{zhang2023building,kim2025multi}.
Large Language Models (LLMs), with extensive encoding of world knowledge and common sense, offer a promising pathway to endow agents with such advanced planning skills \cite{ji2025robobrain,fang2025robix}. 
As shown in Figure \ref{fig:intro:llm_as_planner} (a), at each step, a prompt composed of the current memory (including the status of the agent, status of environment objects, records of past robot actions and tasks, and etc.) and the instruction will be fed to the planner to guide the next action decision \cite{lei2025robomemory,gong2024mindagent}.
Among them, the structured memory provides a global viewpoint to avoid repeated observation and exploration, which is instrumental in improving the task completion efficiency of the agent \cite{tan2025roboos,wang2025karma}.

However, in contrast to the relatively short generated action (less than 10 tokens), the comprehensive memory can result in extremely long prompts, which can be up to tens of thousands of tokens \cite{yang2025efficientnav,yu2025memagent}. 
As shown in Figure \ref{fig:intro:llm_as_planner} (b) and (c), although more retrieved memory segments \footnote{Following \cite{choi2024lota,ji2025robobrain,lei2025robomemory,zhang2023building}, embodied memory is highly structured and can be divided into memory segments. As shown in Figure \ref{fig:intro:llm_as_planner}(a), according to memory category, a memory segment can be the status of one environment object, the recording of one history task, and so on.} can provide more information to the LLM and lead to a higher success rate (SR), it also leads to substantial prefilling time, which becomes the dominant factor in planning latency. 
To accelerate this process, a key observation is that the same memory segments are frequently accessed across different queries and planning steps \cite{yao2025cacheblend,wan2025reca,chen2025retroinfer}.
For instance, the location and state of a foundational object like a table might be relevant to multiple actions, such as placing a cup on it or cleaning it. 
In LLM inference domain, a common optimization is to cache the Key-Value (KV) pairs upon their first computation \cite{gim2024prompt,liu2023cachegen}.
In subsequent reasoning steps, the system can directly retrieve these pre-computed KV pairs for the memory blocks.
This is widely used in applications such as Retrieval-Augmented Generation (RAG) to avoid redundant computation \cite{yao2025cacheblend,jin2024ragcache}.

Through promising, this caching paradigm faces a fundamental challenge when applied to embodied planning.
Unlike the typical LLM inference, which usually processes static text blocks, the various forms of memory in embodied scenarios are often highly dynamic and frequently updated \cite{lei2025robomemory,wang2025karma}. 
For instance, after a robot executes the action ``pick milk from the table'', the memory related to the milk changes, and the state of the table also changes (it no longer holds the milk). 
After each action, only the prefix before the updated memory segment can be reused, while the cache for all subsequent tokens contains invalid information and must be recalculated \cite{jin2024ragcache,yao2025cacheblend}. 
This severely undermines the acceleration benefits. 
To this end, this paper tackles a core problem: within embodied scenarios, how to construct and manage memory efficiently in the face of its frequent updates, thereby improving planning efficiency.

To address these challenges, we propose a KV-Cache-Centric memory management system for efficient embodied planning. 
First, in terms of memory construction, we devise a Static-dynamic Memory Management scheme to mitigate KV cache invalidation caused by memory updates. 
This mechanism groups memory segments by their changing frequency and performs KV computation for different groups at different granularities. 
Second, to reduce the impact of cross-attention ignorance between different groups, we propose Multi-hop Memory Re-computation to recover the interactions between important memories.
Different from previous methods that recompute tokens at specific positions, this technique dynamically selects important memory segments for re-computation according to current query and environmental context. 
Third, we find that under KV recomputation paradigm, the cost of KV loading and recomputation is highly imbalanced across LLM layers, creating pipeline bubbles in traditional loading-computation-parallel schedules. To tackle this, we design a Layer-balanced Memory Loading strategy that orchestrates the workload across layers, effectively minimizing idle time and maximizing hardware utilization.  Our main contributions can be summarized as follows:
\begin{itemize}
   \item We provide a detailed analysis of the fundamental differences in memory construction and management between embodied planning and conventional LLM inference.
   \item We propose static-dynamic memory construction to minimize KV cache invalidation, a multi-hop memory re-computation mechanism to preserve memory interconnection, and a layer-balanced loading scheduler to optimize hardware utilization.
   \item Extensive experiments show that KEEP achieves 2.68$\times$ speedup over text-based memory methods on ALFRED benchmark. Compared with KV re-computation method CacheBlend, KEEP also achieves 4.13\% SR improvement and 1.90$\times$ time-to-first-token (TTFT) reduction.
\end{itemize}

\section{Background}
\label{sec:background}

The design of memory is pivotal for enabling agents to learn from past experiences and improve planning performance \cite{xu2025mem,lei2025robomemory}. Existing memory construction approaches can be broadly categorized into three paradigms: 
\textbf{i) Parametric Memory} involves fine-tuning models on data collected from specific tasks or environments to embed memory directly into the model weights \cite{yao2023retroformer}. But it incurs significant training overhead and often leads to catastrophic forgetting and generalization loss \cite{zhang2025memgen}.
\textbf{ii) Context Memory} stores memory directly as textual information \cite{lei2025robomemory,wang2025karma}. Although straightforward and training-free, the long context length results in substantial computational redundancy during prefilling \cite{liu2025freekv,chen2025retroinfer}.
\textbf{iii) Latent Memory} compresses memory into a fixed-size latent state, such as representative summary tokens \cite{zhang2025memgen,yu2025memagent,wei2025lightmamba}. However, the limited representational capacity often leads to inevitable loss of details and premature forgetting of critical past events \cite{shi2025look}.

In contrast to the above, this work employs a \textbf{KV-Centric Memory}, storing memory as Key-Value (KV) cache pairs of the LLM, which is training-free and maintains memory integrity. However, its \textbf{\textit{scalability}} becomes a challenge: the memory storage grows linearly with the number of memory tokens, which can quickly exceed the capacity of device memory (e.g., GPU HBM) \cite{liu2023cachegen,liu2025chunkkv,hu2025efficient}. To address this, following \cite{liu2023cachegen,yao2025cacheblend,sun2025breaking}, our system utilizes slower but higher-capacity memory (e.g., CPU RAM) for KV storage and only loads KV pairs of relevant memory segments to the GPU in each inference.
For efficient KV loading, we orchestrate a layer-grained pipeline to overlap KV loading and computation, thereby mitigating the latency cost. We will further discuss this in Section \ref{sec:method3:loading}.


\section{When KV Reuse Meets Embodied Planning}
\label{sec:motivation}

As established in Section \ref{sec:intro} and \ref{sec:background}, a common optimization for reusable memory is to cache its KV to avoid repeated computation. 
Existing KV reuse methods can be categorized as follows. 
\textbf{i) Prefix Caching} reuses the KV cache only for the identical prefix of the LLM input, which guarantees output quality but requires an exact match of the prefix \cite{zheng2023efficiently,kwon2023efficient}. 
For scenarios like RAG, where memory selection varies with different queries (which is also the case in embodied planning), \textbf{ii) Full KV Reuse} segments long context into fixed-length blocks and calculates their KV states individually, enabling more flexible KV reuse \cite{gim2024prompt,jin2024ragcache}.
To address the loss of cross-attention between these independently cached blocks, \textbf{iii) KV Recomputation methods} identify a small ratio of important tokens in model inference and selectively recompute their KV cache in subsequent layers \cite{yao2025cacheblend,hu2024epic,cao2025sparse}, thus increasing model accuracy with negligible extra computation cost.
While these methods have shown great success in general LLM inference, they face two \textbf{\textit{stability}} challenges in embodied planning.

\begin{figure}[!tb]
    \centering
    \includegraphics[width=1.0\linewidth]{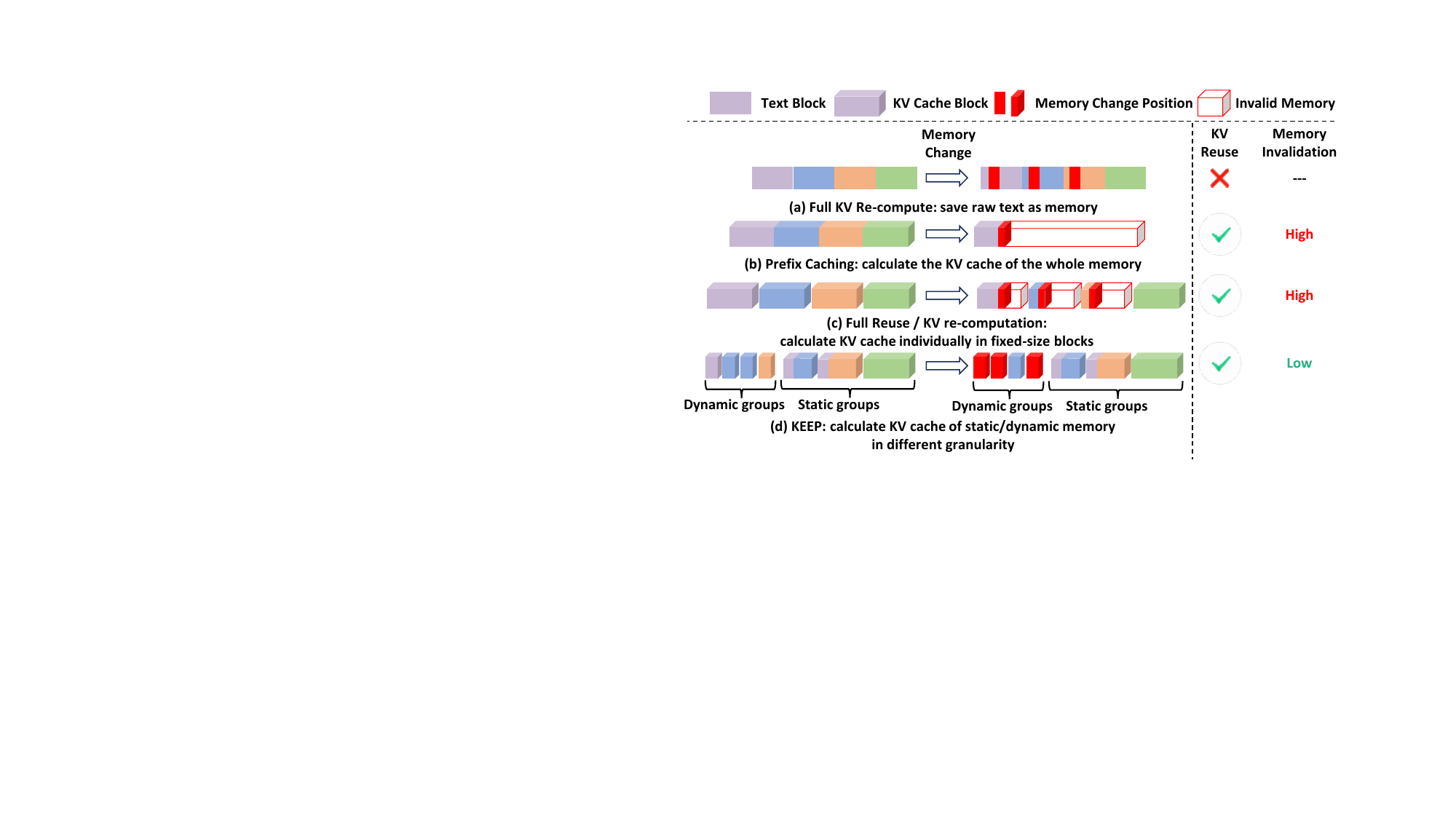}
    \caption{Comparison with previous KV reuse methods on memory construction.}
    \label{fig:challeng1:construct}
    \vspace{-7pt}
\end{figure}

\begin{figure}[!tb]
    \centering
    \includegraphics[width=1.0\linewidth]{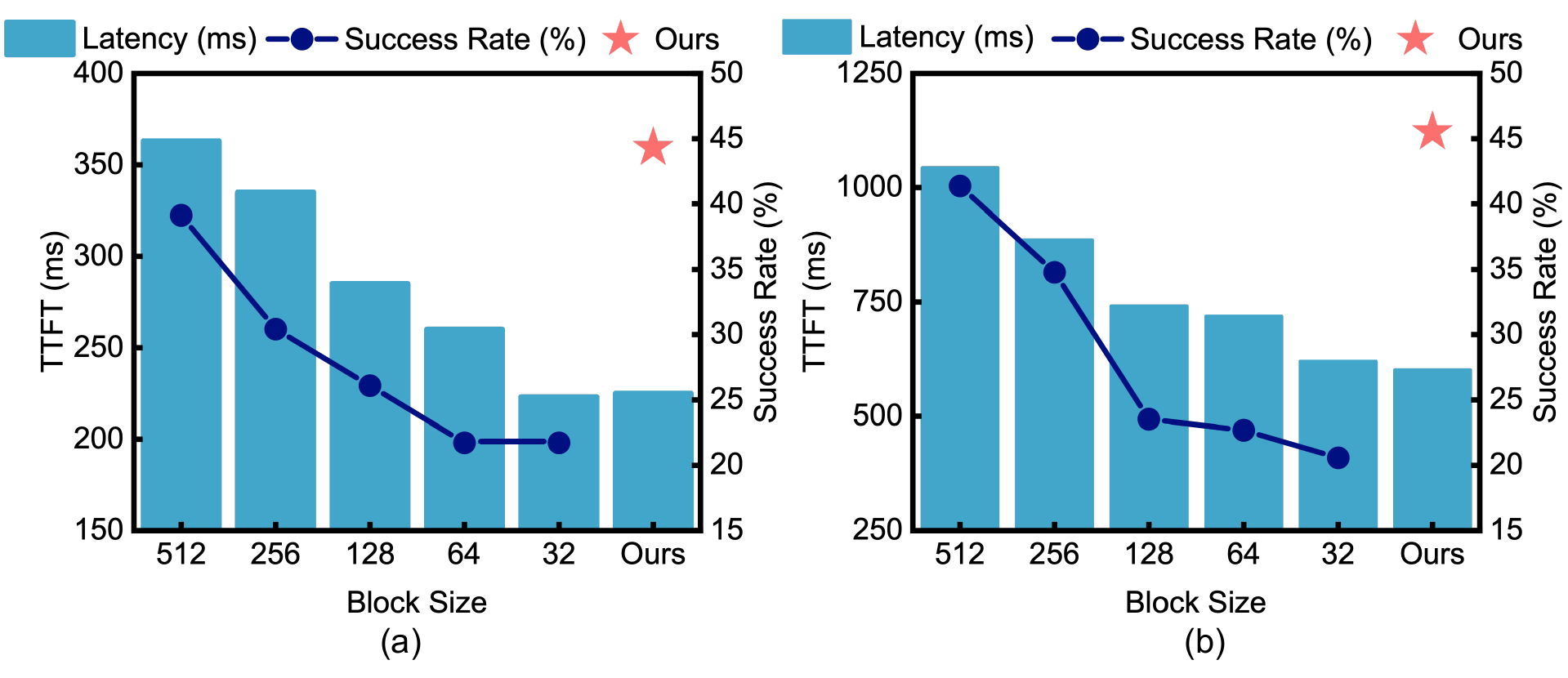}
    \caption{Impact of different block sizes for KV recomputation methods, evaluated on ALFRED with CacheBlend method using (a) Qwen-14B and (b) Qwen-32B (INT4).
    }
    \label{fig:challeng1:latency_SR_blocksize}
\end{figure}

\textbf{Challenge 1: For memory construction, the coarse-grained, fixed-sized block partitioning fails to align with the fine-grained and rapid updating embodied memory.}
As shown in Figure \ref{fig:challeng1:construct}, for prefix caching, a single memory update will make the whole following memory invalid.
For full reuse or KV recomputation methods, which use fixed-size coarse-grained blocks to save KV, for a minor update, the KV after the update position in the same block must be invalidated and recomputed. 
While decreasing the block size could reduce this invalidation, it severs the cross-attention ignorance between different blocks, leading to a drop in planning accuracy, which is shown in Figure \ref{fig:challeng1:latency_SR_blocksize}. So a trade-off between memory management granularity and memory interconnection preservation is needed.

\begin{figure}[!tb]
    \centering
    \includegraphics[width=1.0\linewidth]{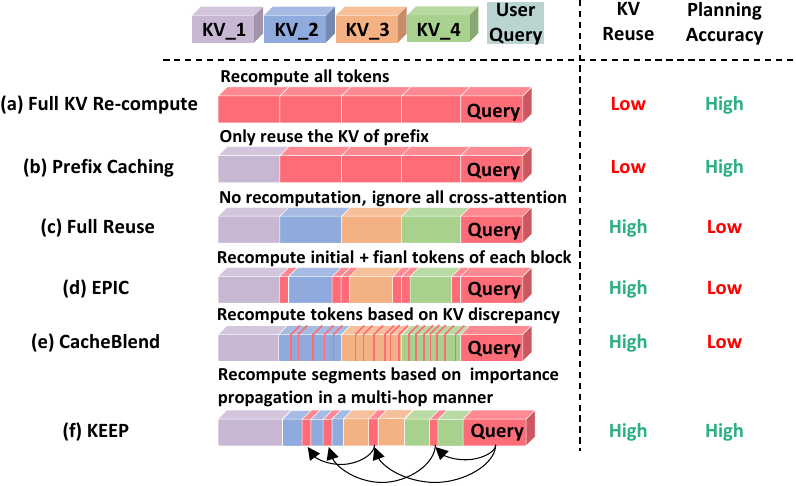}
    \caption{Method comparison on KV recomputation.}
    \label{fig:challenge2:recomputation}
    \vspace{-10pt}
\end{figure}

\textbf{Challenge 2: For KV recomputation, the static recomputation paradigm of existing methods fails to adapt to the dynamic and context-dependent memory importance in embodied scenarios.}
As shown in Figure \ref{fig:challenge2:recomputation}, in planning process, full KV re-computation and prefix caching show low KV reuse, leading to repeated computation and high TTFT.
Full reuse methods \cite{gim2024prompt} ignores all cross-attention between groups, leading to low accuracy.
KV recomputation approaches like EPIC \cite{hu2024epic}, heuristically recompute tokens at fixed positions (e.g., the start and end of a block). And CacheBlend \cite{yao2025cacheblend} selects tokens based on the discrepancy between the full-attention KV state and the cached KV state, in which the recomputation token is only decided by its prefix.
However, in embodied planning, the memory relevance is highly dependent on the query and the whole memory context.
For instance, a memory segment \textit{\{``object'':``key'', ``position'':``on the table'', ...\}} is critically important for the task ``unlock the door'' but only marginally relevant for the task ``find the door''.
A static KV recomputation mechanism might overlook important memory interconnections or waste computation regardless of the memory relevance in current situation.
So a dynamic recomputation strategy that actively chooses important memory interconnections is essential for achieving both high accuracy and efficiency. We will further discuss this in Section \ref{sec:method2:multi-hop}.

\textbf{Overview.}
In this work, we propose a KV-centric memory management system for efficient embodied planning. To minimize KV cache invalidation while preserving interconnections, we propose Static-Dynamic Memory Construction (Section \ref{sec:method1:static-dynamic}) to group memory by update frequency and manage different groups with different granularities. To recover critical memory interactions, we propose Multi-hop Memory Re-computation (Section \ref{sec:method2:multi-hop}) to conduct memory recomputation based on the query and current context. For the large KV cache footprint, we propose a hierarchical storage strategy with Layer-balanced Memory Loading (Section \ref{sec:method3:loading}) that leverages a pipelined schedule for better scalability and efficiency.


\vspace{-5pt}

\section{Method}
\label{sec:Method}

\begin{figure}[!tb]
    \centering
    \includegraphics[width=1.0\linewidth]{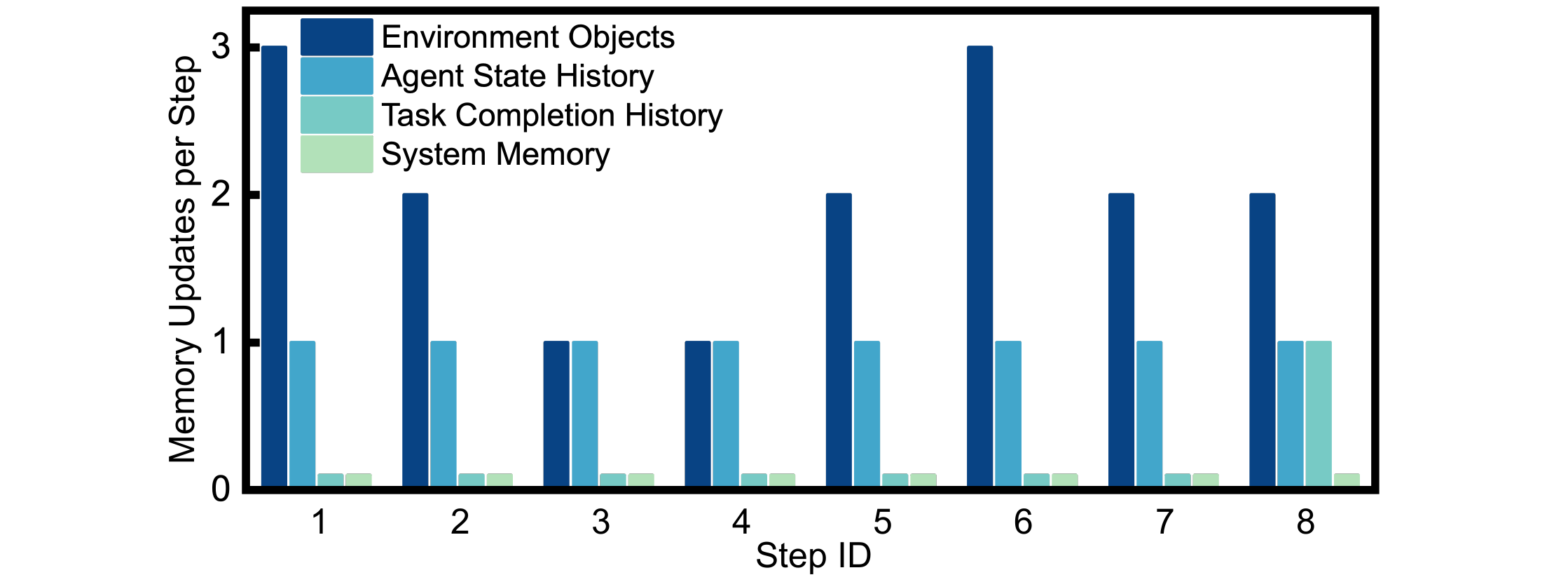}
    \caption{Different memories show different update frequencies. Here we use a coarse-grained memory classification as an example.}
    \label{fig:method1:memory_change}
    \vspace{-10pt}
\end{figure}

\subsection{Static-dynamic Memory Construction}
\label{sec:method1:static-dynamic}
\textbf{Motivation.} As discussed in Section \ref{sec:motivation},  existing KV reuse methods split long context into fixed-length blocks to construct memory.
In embodied planning, with different block sizes, this strategy either shows high re-computation costs because of memory updates or shows planning accuracy drops because of severe cross-attention ignorance. 
However, we find that the update frequency of different embodied memories actually varies a lot.
As shown in Figure \ref{fig:method1:memory_change}, memories such as the status of environmental objects usually show a high updating frequency. And memories such as completed tasks change rarely, as it is just a record of the agent's working history.
In addition, even the memory segments within the same cluster have various changing frequencies.
For example, the changing frequency of the position of a table is much lower than the position of a book.
So, for different memory, using fixed memory management granularity is sub-optimal.

Based on this, we propose the Static-Dynamic Memory Construction mechanism. Its core principle is to manage memory based on changing frequency: highly dynamic memory segments are calculated separately to prevent unnecessary KV cache invalidation, while relatively static memories are grouped and calculated KV together to preserve their intrinsic contextual relationships.
To be specific, we employ a sentence encoder to cluster memory into groups based on semantics \footnote{Here we use sentence-transformers-mpnet-base-v2 \cite{song2020mpnet} as the sentence encoder. Actually, any widely used sentence encoder is reasonable here.}.
A group is classified as static if all its memory segments have remained unchanged for the recent $t$ steps. For these groups, the KV cache is computed with full cross-attention between segments within the group, thereby preserving their rich interconnections. 
Conversely, a group is labeled as dynamic if any of its memory has been updated within the last $t$ steps. The KV cache for a dynamic group is computed individually per segment, preventing a single update from invalidating the cache of the entire group.
A static group will transition to a dynamic state if any of its members are updated. Conversely, a dynamic group that remains stable for $t$ steps will be changed to a static group. When a group's category changes, we recompute its KV cache according to the new computation paradigm.
This will not introduce much cost, as it happens at a low frequency (usually once in tens of steps). In practice, we set $t=10$, which is a typical length of one task.

To adapt to the different KV computation mechanisms for different groups, during the retrieval phase for a given query, we also apply different strategies. For static groups, retrieval is performed at the group level; the entire group is selected as an interconnected unit. For dynamic groups, it is performed at the memory segment level for better flexibility. 

This hybrid strategy effectively avoids the huge KV invalidation caused by dynamic memory and the memory interconnection ignorance between memory segments in static groups.
However, the interconnection between dynamic memory segments and different groups is still not considered. We solve this by KV re-computation on important memories, further discussed in Section \ref{sec:method2:multi-hop}.

\vspace{-3pt}
\subsection{Multi-hop Memory Re-computation}
\label{sec:method2:multi-hop}
\begin{figure}[!tb]
    \centering
    \includegraphics[width=1.0\linewidth]{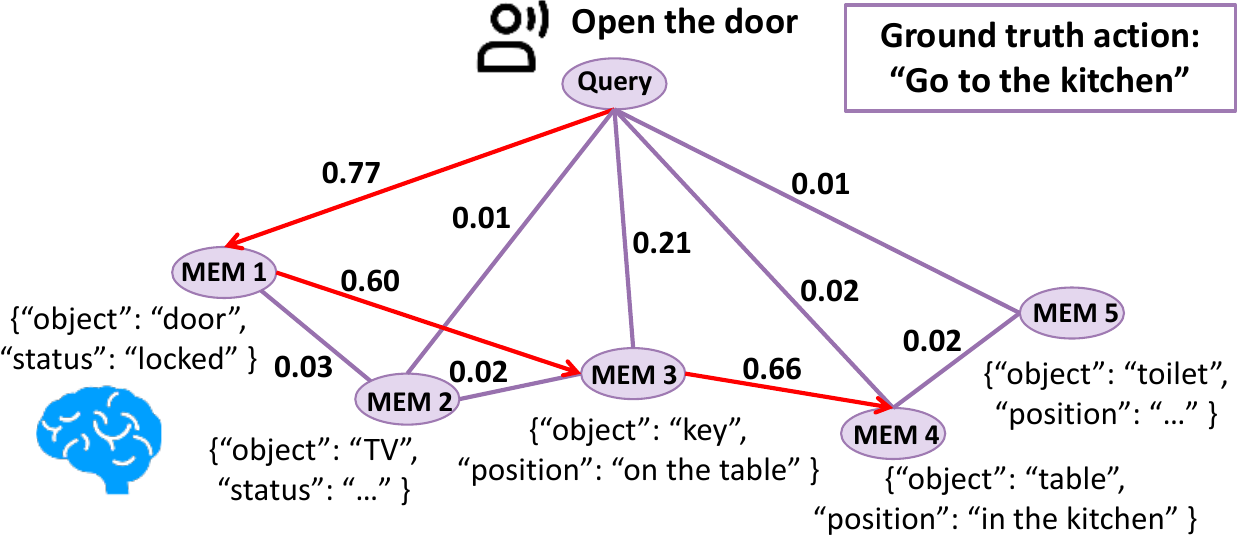}
    \caption{The memory importance evaluation works in a multi-hop manner. Here we show an example of importance propagation with five memory segments. The number above the memory connection is the normalized cross-attention score between two memory segments (some minor attention score is not shown). The red arrows show that we find important memories in a multi-hop manner.}
    \label{fig:method2:multi_hop}
    \vspace{-10pt}
\end{figure}

\textbf{Motivation.} As discussed in Section \ref{sec:motivation}, existing KV reuse methods use static paradigms to select important memories for recomputation. 
However, in embodied planning, a single memory segment's importance often depends on the query and memory context.
For example, as shown in Figure \ref{fig:method2:multi_hop}, the important memory segment of ``table'' is hard to find unless the logic chain \textit{``Locked Door -> Key -> Table''} is fully used.
So, a multi-hop evaluation mechanism is necessary for accurately quantifying memory importance.

Based on this, we propose Multi-hop Memory Re-computation to identify and recompute critical memories. Following prior work \cite{yao2025cacheblend}, we utilize attention distributions to evaluate memory importance. However, unlike methods that study the attention distribution at fixed positions \cite{hu2024epic}.
Our approach evaluates memory importance using importance propagation. As shown in figure \ref{fig:method2:multi_hop}, according to attention distribution, it first identifies memories important to the query, then uses these memories to find secondary important memories critical to them, and iterates this process to uncover crucial inter-memory relationships.
To be specific, after obtaining the attention scores at layer $i$, we proceed as follows:
\begin{itemize}
\item \textbf{Initialization}: The average attention scores between the query and all memories to be evaluated are used as the initial importance scores for the memories.
\item \textbf{Importance Propagation}: A ratio of top $r_i$ memories based on current importance scores is selected as the relevant set. 
The importance scores of all memories are then updated by averaging the cross-attention weights from the current relevant memory set.
This step effectively allows important memories to ``propagate'' their importance to other memories critical to them.
\item \textbf{Convergence Check}: Steps of importance propagation are repeated until the composition of the relevant memory set stabilizes between iterations.
\item \textbf{Selective Recomputation}: In layer $i+1$, only KV cache for memories in the final stabilized relevant set is recomputed.
\end{itemize}
Following CacheBlend \cite{yao2025cacheblend}, we recompute all memories in the first layer and gradually reduce the recomputation ratio $r_i$ of subsequent layers. Finally, we maintain a relatively low average recomputation ratio $r$ while preserving accuracy by focusing recomputation on the most critical memory discovered through multi-hop importance propagation.
We will discuss the impact of $r$ in the ablation study.

\textbf{System advantage}: Another key difference between our strategy and existing kv recomputation methods \cite{jin2024ragcache} is on the recomputation granularity.
We evaluate memory importance and conduct recomputation in memory segment granularity, rather than token granularity \cite{yao2025cacheblend}.
This enables efficient, contiguous loading of only the non-recomputed KV blocks. In contrast, token-granularity methods must load the entire cache and perform in-place updates for recomputed tokens to preserve storage continuity, incurring significantly higher I/O overhead.
Crucially, our method maintains accuracy despite the coarser granularity by accurately identifying critical memory by importance propagation.
And the propagation process will not introduce much latency cost, and it can be calculated in parallel with the MLP of the same layer.

\vspace{-3pt}

\subsection{Layer-balanced Memory Loading}
\label{sec:method3:loading}
\begin{figure}[!tb]
    \centering
    \includegraphics[width=0.9\linewidth]{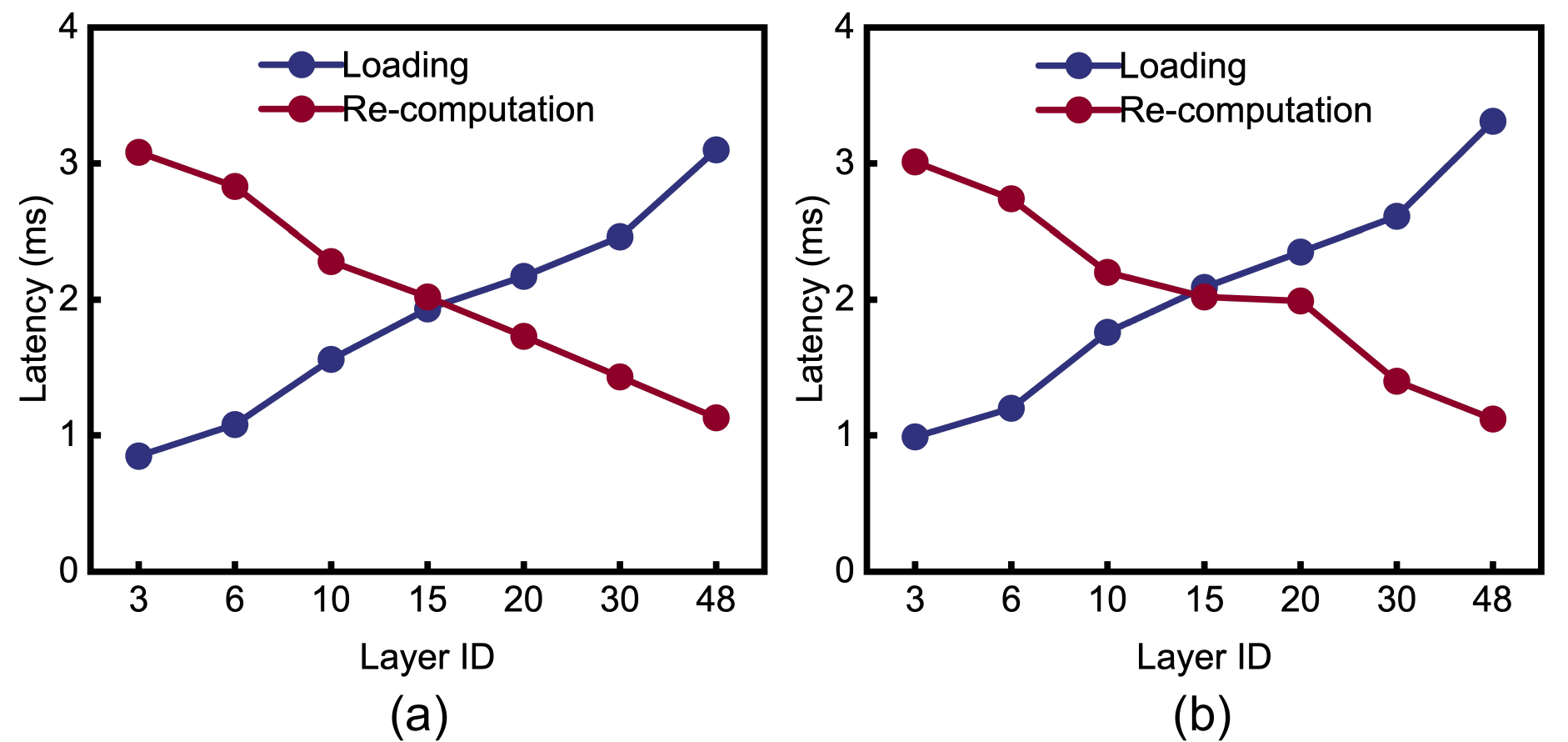}
    \caption{Unbalanced KV loading and computation cost of different layers for (a) Qwen-14B and (b) Qwen-32B (INT4).
    }
    \label{fig:method3:latency_layerID}
\end{figure}

\begin{figure}[!tb]
    \centering
    \includegraphics[width=1.0\linewidth]{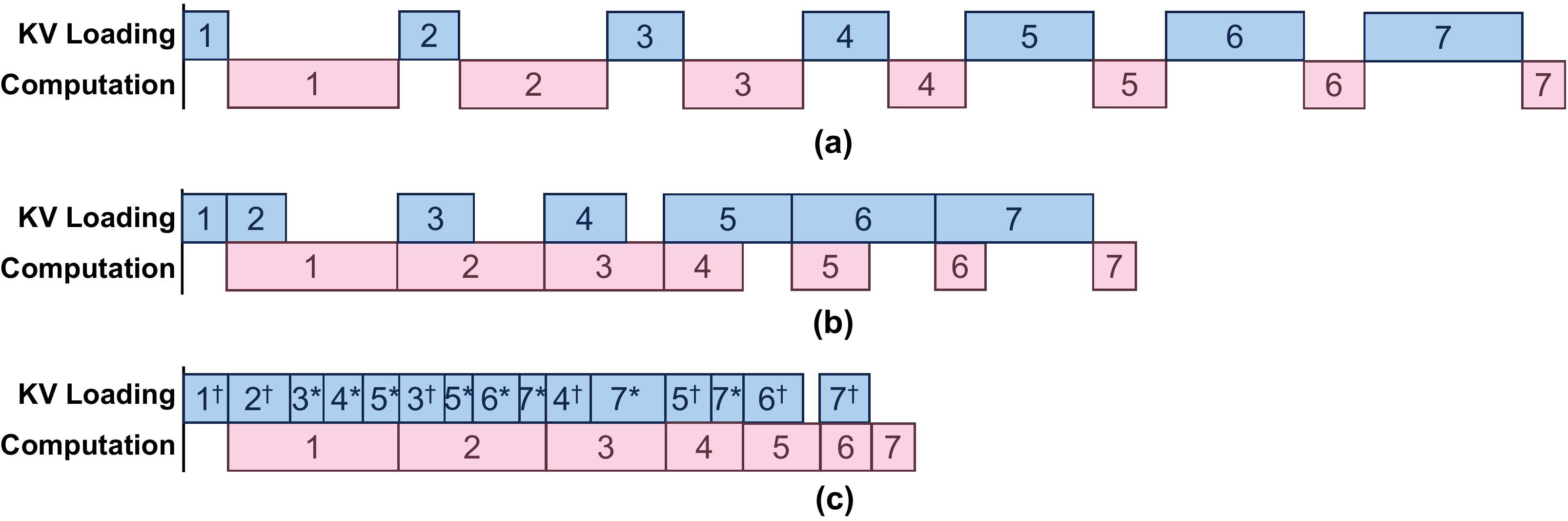}
    \caption{Pipeline comparison between (a) separate KV loading and computation, (b) computation parallel with KV loading of the next layer, and (c) layer-balanced memory loading. Here we use a 7-layer model as an example. $n^*$ means load the memory which is guaranteed not to be recomputed in layer $n$. $n^\dagger$ means finish loading the KV needed in layer $n$.
    }
    \label{fig:method3:pipeline}
    \vspace{-5pt}
\end{figure}

\textbf{Motivation.} As discussed in Section \ref{sec:background}, 
we store the KV cache memory on slower but higher-capacity memory (e.g., CPU RAM) to solve the scalability problem. However, this will introduce extra KV loading cost before the attention computation of each layer.
To hide this cost, a common technique \cite{pan2025kvflow,sun2024shadowkv,liu2023cachegen} is to parallelize the computation of the current layer with the KV loading for the next layer \footnote{In fact, when combined with KV re-computation, we can not know the KV to load next layer before KV importance evaluation of this layer. Here we just load the memory of the next layer according to current layer, as the memory to recompute will not change much between adjacent layers. However, for later layers, the memory needed to load will be more different.}.
However, we find that this strategy becomes inefficient when combined with our recomputation mechanism. As shown in Figure \ref{fig:method3:latency_layerID}, the workload across layers is highly imbalanced: earlier layers, which recompute a larger proportion of memories, have less KV data to load. Conversely, later layers, with a lower recomputation ratio, must load a much larger portion of KV cache. 
As illustrated in Figure \ref{fig:method3:pipeline} (b), this imbalance creates pipeline bubbles both in KV loading for early layers and computation in later layers.
Consequently, the inherent imbalance prevents the standard pipeline from fully amortizing the loading overhead.

To address this, we propose Layer-balanced Memory Loading. Our approach is grounded in a key observation: for any given memory segment, if it is not recomputed in an earlier layer, it will not be recomputed in any subsequent layers. This is because the input hidden state corresponding to that memory is discarded after the not-to-recompute decision. This makes it impossible to generate new KV cache for this memory in later layers. Consequently, KV cache for such memories must be loaded for all subsequent layers.

Leveraging this, our strategy proactively manages the KV loading pipeline, which is shown in Figure \ref{fig:method3:pipeline} (c). In earlier layers, if the KV loading for the next layer finishes before the current layer's computation, the idle loading engine immediately begins pre-loading the KV cache for memories guaranteed not to be recomputed in future layers. 
This pre-loading continues until current layer's computation finishes.
This strategy both eliminates the loading bubble in the early stages and reduces the KV loading burden for later layers, thus also eliminating the computation bubble for later layers.
This cross-layer, proactive pre-loading mechanism effectively redistributes the loading overhead from later layers to underutilized earlier layers, resulting in a more balanced execution pipeline.

\section{Implementation}

We implement KEEP on top of vLLM \cite{kwon2023efficient} with about 2.8K lines of code in Python based on PyTorch v2.0. We integrate our method into the LLM serving engine through three interfaces:
\begin{itemize}
\item\texttt{load\_memory(memory\_id, layer\_id) -> KV Cache:} given a memory id, KEEP loads its KV cache of an indicated layer.
\item\texttt{prefill\_layer(input, layer\_id, KV cache) -> output, updated KV cache:} KEEP takes the input and KV cache for an indicated layer to recompute KV of important memories. Then it outputs the input hidden state for the next layer and the updated KV cache for decoding process.
\item\texttt{importance\_evaluation(layer\_id, memory\_ids, attention) -> memory\_ids:} KEEP uses the attention scores to identify the important memories to be re-computed in the next layer.
\end{itemize}
For \texttt{load\_memory}, we use a hash table to record the storage place of each memory. If a needed memory is not in GPU, we use \texttt{torch.cuda()} to load it from CPU RAM. 
In model inference, three threads are used to pipeline the computation (\texttt{prefill\_layer}) of layer $i$, the memory importance evaluation (\texttt{importance\_evaluation}) of layer $i$, and the KV cache loading (\texttt{load\_memory}) of layer $i+1$ and the not-recomputed memories in future layers. 
We also call a \texttt{synchronize()} function to ensure two dependencies: 1) the memory importance evaluation for layer $i$ and memory loading for layer $i+1$ should finish before the computation of layer $i+1$; 2) the attention computation of layer $i$ should finish before the memory importance evaluation of layer $i$.


\vspace{-5pt}
\section{Experiments}
\label{sec:exp}


\begin{table}[!tb]
    \centering
    \caption{Evaluation on ALFRED dataset.}
    \label{tab:acc:alfred}
    \scalebox{0.88}{
    \begin{tabular}{lccccc}
    \toprule
    \textbf{Methods}  & \textbf{Model} & \textbf{SR} & \textbf{TTFT} \\
    \midrule
    LLM-Planner \cite{song2023llm}  & GPT-3 & 16.45\% & - \\
    FLARE \cite{kim2025multi} & GPT-4 & 40.05\% & - \\
    KARMA \cite{wang2025karma} & GPT-4o & 43.00\% & - \\
    LoTA (Full Recompute) \cite{choi2024lota}& Qwen-2.5-14B & 44.63\% & 0.410s \\
    LoTA + Full Reuse \cite{gim2024prompt}& Qwen-2.5-14B & 34.41\% & 0.210s \\
    LoTA + CacheBlend \cite{yao2025cacheblend}& Qwen-2.5-14B & 39.36\% & 0.363s \\
    LoTA + KEEP  & Qwen-2.5-14B & 44.30\% & 0.236s \\
    \midrule
    LoTA (Full Recompute) \cite{choi2024lota} & Qwen-2.5-32B (INT4) & 45.81\% & 1.213s \\
    LoTA + Full Reuse \cite{gim2024prompt} & Qwen-2.5-32B (INT4) & 35.72\% & 0.602s \\
    LoTA + CacheBlend \cite{yao2025cacheblend}& Qwen-2.5-32B (INT4) & 41.37\% & 1.209s \\
    LoTA + KEEP & Qwen-2.5-32B (INT4) & 45.50\% & 0.635s \\
    
        \bottomrule
    \end{tabular}
    }
    \vspace{-7pt}
\end{table}

\begin{table}[!tb]
    \centering
    \caption{Evaluation on WAH-NL dataset.}
    \label{tab:acc:wah}
    \scalebox{0.88}{
    \begin{tabular}{lccccc}
    \toprule
\textbf{Methods} & \textbf{Model} & \textbf{Sub-SR} & \textbf{TTFT} \\
\midrule
LoTA (Full Recompute) \cite{choi2024lota}  & Qwen-2.5-14B & 19.58\% & 0.362s \\
LoTA + Full Reuse \cite{gim2024prompt} & Qwen-2.5-14B & 9.08\% & 0.203s \\
LoTA + CacheBlend \cite{yao2025cacheblend}& Qwen-2.5-14B & 16.21\% & 0.341s \\
LoTA + KEEP & Qwen-2.5-14B & 19.52\% & 0.226s \\
\midrule
LoTA (Full Recompute) \cite{choi2024lota} & Qwen-2.5-32B (INT4) & 20.48\% & 1.221s \\
LoTA + Full Reuse \cite{gim2024prompt} &  Qwen-2.5-32B (INT4)& 12.74\% & 0.578s \\
LoTA + CacheBlend \cite{yao2025cacheblend}&  Qwen-2.5-32B (INT4)& 16.88\% & 1.148s \\
LoTA + KEEP &  Qwen-2.5-32B (INT4)& 20.25\% & 0.601s \\
        \bottomrule
    \end{tabular}
    }
    \vspace{-10pt}
\end{table}

\subsection{Experiment Setup}
We evaluate KEEP on ALFRED \cite{shridhar2020alfred} and WAH-NL \cite{choi2024lota} benchmark, using the Qwen-2.5 model family \cite{team2024qwen2}. 
In each step, the agent can access the object states in the environment, the agent state, and the task history \cite{choi2024lota}. 
Then the agent updates its memory and uses LLM planner to predict the next action. 
For action completion, we follow \cite{choi2024lota} and \cite{yang2025efficientnav}, using the agent action APIs embedded in the platform.
Following \cite{shridhar2020alfred} and \cite{choi2024lota}, we evaluate planning accuracy using Success Rate (SR) and Subgoal success rate (Sub-SR), which means the ratio of tasks/sub-tasks that the agent can complete. For planning efficiency, we evaluate TTFT, as prefilling time accounts for over 90\% of the planning latency. 
Our memory retrieval strategy follows \cite{choi2024lota}.
Each model is deployed on one NVIDIA A6000 GPU. For Qwen-32B model, we use INT4 weight-only quantization \cite{lin2024awq}.

\subsection{Main Results}

Table \ref{tab:acc:alfred} and \ref{tab:acc:wah} show the evaluation results on ALFRED and WAH-NL benchmark. On ALFRED, compared with traditional methods FLARE \cite{kim2025multi} and KARMA \cite{wang2025karma}, KEEP shows 4.25\% SR and 1.30\% SR improvement with a much smaller LLM planner, by timely recording and updating memory in each action step.
Compared with LOTA (full recompute) \cite{choi2024lota}, KEEP shows 1.74$\times$ and 1.91$\times$ TTFT reduction using Qwen-14B and Qwen-32B, with negligible accuracy loss.
Compared with the full reuse method \cite{gim2024prompt} and CacheBlend, KEEP shows 9.89\% and 4.94\% SR improvement on 14B model and 9.78\% and 4.13\% on 32B model, by using importance propagation to recover cross-attention between important memories. 
Compared with CacheBlend, KEEP also shows 1.54$\times$ and 1.90$\times$ TTFT reduction on 14B and 32B models, by using static-dynamic memory construction to avoid KV invalidation.
On WAH-NL, compared with CacheBlend, KEEP also shows 3.31\% and 3.37\% Sub-SR improvement, and 1.51$\times$ and 1.91$\times$ TTFT reduction.

\begin{figure}[!tb]
    \centering
    \includegraphics[width=0.88\linewidth]{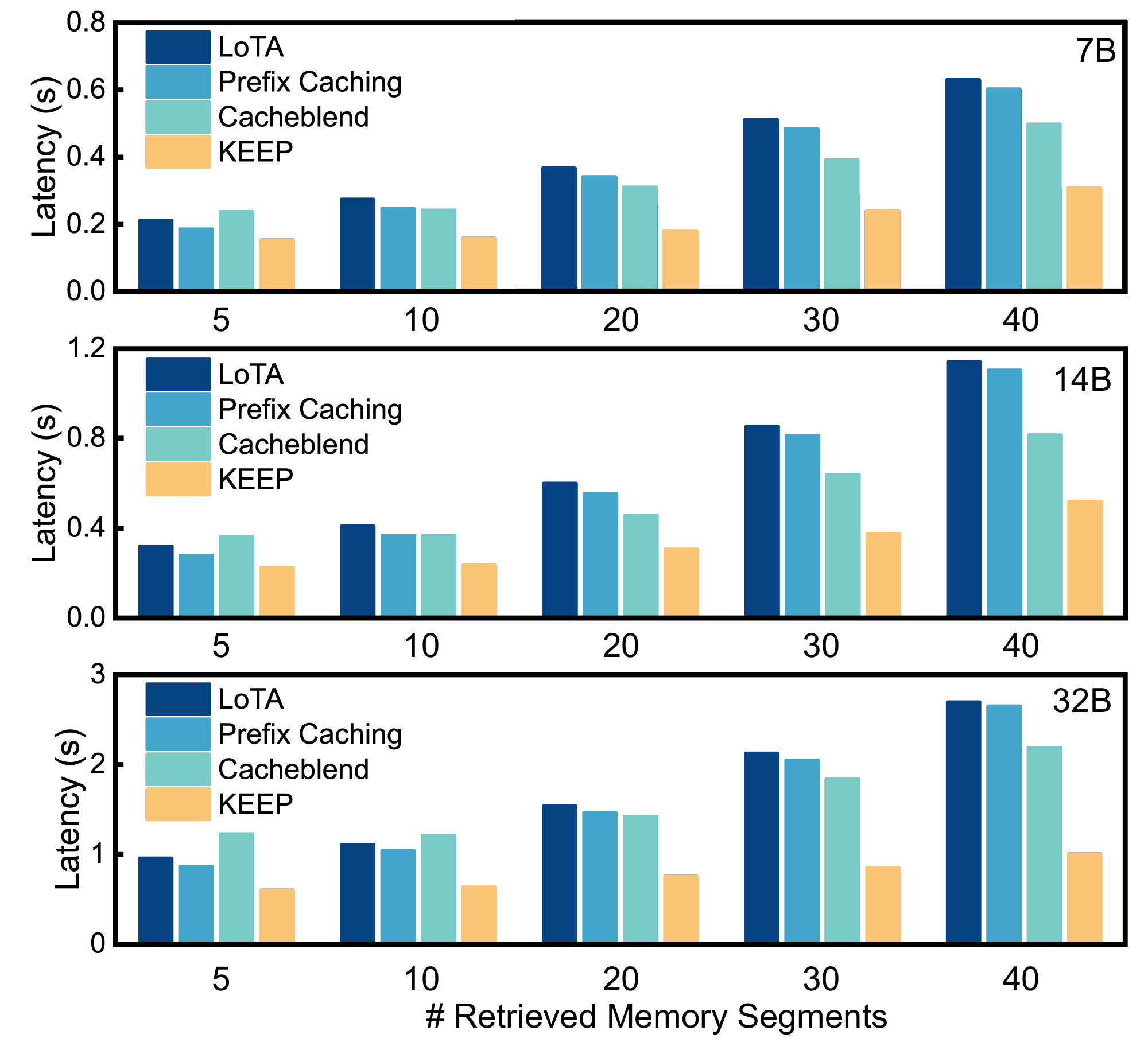}
    \caption{Latency comparison with different numbers of retrieved memory segments.}
    \label{fig:exp:latency}
    \vspace{-5pt}
\end{figure}

\textbf{Latency comparison with different numbers of retrieved memory segments.}
As discussed in Section \ref{sec:intro}, more retrieved memory segments can provide more information to the LLM planner, while it will also introduce a longer prompt and higher latency.
Here we evaluate how TTFT scales with the number of retrieved memory segments on ALFRED benchmark.
As shown in Figure \ref{fig:exp:latency}, the latency of full computation and prefix caching increases rapidly with segments number.
The latency of CacheBlend also grows as more memory segments contain more memory updates, causing more KV invalidation.
However, the TTFT of KEEP increases slower than other methods, as static-dynamic memory construction can reduce memory invalidation.
With 40 segments, KEEP shows 2.19$\times$ and 2.68$\times$ TTFT reduction over full recomputation, and 1.56$\times$ and 2.17$\times$ TTFT reduction over CacheBlend, on 14B and 32B model.

\subsection{Ablation Study}
\textbf{Individual influence of our methods.}
The individual influence of our methods is shown in Table \ref{tab:ablation_study}, evaluated on ALFRED. Missing static-dynamic memory construction will introduce a 6.94\% SR drop, because of ignoring cross-attention between static memory segments.
It will also introduce 1.54$\times$ TTFT increase, as managing KV in fixed-size blocks will introduce high KV invalidation, which needs to be re-computed.
Missing multi-hop memory re-computation introduces a 2.52\% SR drop, as the cross-attention between importance memory can not be fully recovered without importance propagation.
Missing layer-balanced memory loading introduces 1.20$\times$ TTFT increase, which comes from memory loading bubbles in early layers and computation bubbles in latter layers.

\begin{table}[!tb]
\centering
\caption{Individual influence of our methods.}
\scalebox{0.9}{
\begin{tabular}{lcc}
\toprule
\textbf{Methods} & \textbf{SR} & \textbf{TTFT} \\
\midrule
KEEP & 44.30\% & 0.230s \\
w/o static-dynamic memory construction & 37.36\% & 0.355s \\
w/o multi-hop memory re-computation & 41.78\% & 0.228s \\
w/o layer-balanced memory loading & 44.30\% & 0.273s \\
\bottomrule
\end{tabular}
}
\label{tab:ablation_study}
\vspace{-3pt}
\end{table}

\begin{figure}[!tb]
    \centering
    \includegraphics[width=0.9\linewidth]{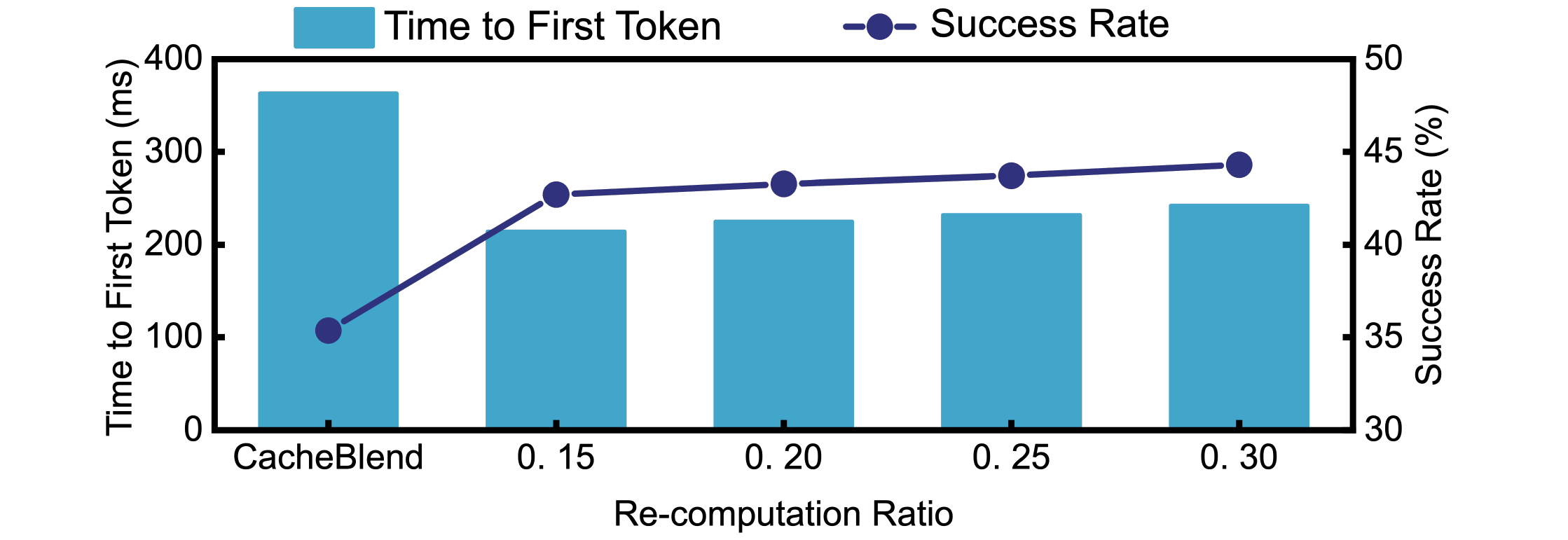}
    \caption{Ablation study on different recomputation ratios.}
    \label{fig:exp:compute_ratio}
    \vspace{-12pt}
\end{figure}

\textbf{Impact of recomputation ratios.}
The individual influence on recomputation ratios is shown in Figure \ref{fig:exp:compute_ratio}, evaluated on ALFRED using Qwen-14B. KEEP shows robust performance across different recomputation ratios.
KEEP maintains high SR even with a low recomputation ratio, as the cross-attention between the most important memories can be recovered by multi-hop memory recomputation.
KEEP maintains a low TTFT for a high recomputation ratio, 
as the latency of recomputation is effectively overlapped with KV loading process, reducing the impact on overall inference speed.

\section{Conclusion}
We propose KEEP, a KV-Cache-centric memory management system for efficient embodied planning. To avoid high memory invalidation caused by memory update in embodied scenarios, we propose static-dynamic memory management to manage dynamic and static memory at different granularities.
To recover cross-attention between important memories, we propose multi-hop memory re-computation to find important memories by importance propagation and recompute their KV cache.
We also propose layer-balanced memory loading to balance the KV loading and computation overhead in different layers.
Compared with CacheBlend, KEEP shows 4.13\% SR improvement and 1.90$\times$ TTFT reduction on ALFRED.

\section{Acknowledgments}
This work was supported in part by New Generation Artificial Intelligence-National Science and Technology Major Project (No. 2025ZD0122502), in part by Beijing Municipal Science and Technology Program under Grant Z241100004224015, in part by NSFC (No. 92464104, 62495102), in part by the National Key Research and Development Program under Grant 2024YFB4505004, and in part by Shenzhen Key Industry R\&D Project (No. ZDCY20250901105036006): Research and Development of High-Efficiency Edge Chips for "Brain-Cerebellum" Coordination in Embodied Intelligence.

\newpage

\bibliographystyle{ACM-Reference-Format}
\bibliography{main}

@article{zhang2023building,
  title={Building cooperative embodied agents modularly with large language models},
  author={Zhang, Hongxin and Du, Weihua and Shan, Jiaming and Zhou, Qinhong and Du, Yilun and Tenenbaum, Joshua B and Shu, Tianmin and Gan, Chuang},
  journal={arXiv preprint arXiv:2307.02485},
  year={2023}
}

@inproceedings{kim2025multi,
  title={Multi-Modal Grounded Planning and Efficient Replanning For Learning Embodied Agents with A Few Examples},
  author={Kim, Taewoong and Kim, Byeonghwi and Choi, Jonghyun},
  booktitle={Proceedings of the AAAI Conference on Artificial Intelligence},
  volume={39},
  number={4},
  pages={4329--4337},
  year={2025}
}

@inproceedings{ji2025robobrain,
  title={Robobrain: A unified brain model for robotic manipulation from abstract to concrete},
  author={Ji, Yuheng and Tan, Huajie and Shi, Jiayu and Hao, Xiaoshuai and Zhang, Yuan and Zhang, Hengyuan and Wang, Pengwei and Zhao, Mengdi and Mu, Yao and An, Pengju and others},
  booktitle={Proceedings of the Computer Vision and Pattern Recognition Conference},
  pages={1724--1734},
  year={2025}
}

@inproceedings{wang2025karma,
  title={Karma: Augmenting embodied ai agents with long-and-short term memory systems},
  author={Wang, Zixuan and Yu, Bo and Zhao, Junzhe and Sun, Wenhao and Hou, Sai and Liang, Shuai and Hu, Xing and Han, Yinhe and Gan, Yiming},
  booktitle={2025 IEEE International Conference on Robotics and Automation (ICRA)},
  pages={1--8},
  year={2025},
  organization={IEEE}
}

@article{lei2025robomemory,
  title={RoboMemory: A Brain-inspired Multi-memory Agentic Framework for Lifelong Learning in Physical Embodied Systems},
  author={Lei, Mingcong and Cai, Honghao and Que, Binbin and Cui, Zezhou and Tan, Liangchen and Hong, Junkun and Hu, Gehan and Zhu, Shuangyu and Wu, Yimou and Jiang, Shaohan and others},
  journal={arXiv e-prints},
  pages={arXiv--2508},
  year={2025}
}

@inproceedings{gong2024mindagent,
  title={Mindagent: Emergent gaming interaction},
  author={Gong, Ran and Huang, Qiuyuan and Ma, Xiaojian and Noda, Yusuke and Durante, Zane and Zheng, Zilong and Terzopoulos, Demetri and Fei-Fei, Li and Gao, Jianfeng and Vo, Hoi},
  booktitle={Findings of the Association for Computational Linguistics: NAACL 2024},
  pages={3154--3183},
  year={2024}
}

@article{fang2025robix,
  title={Robix: A unified model for robot interaction, reasoning and planning},
  author={Fang, Huang and Zhang, Mengxi and Dong, Heng and Li, Wei and Wang, Zixuan and Zhang, Qifeng and Tian, Xueyun and Hu, Yucheng and Li, Hang},
  journal={arXiv preprint arXiv:2509.01106},
  year={2025}
}

@article{tan2025roboos,
  title={Roboos: A hierarchical embodied framework for cross-embodiment and multi-agent collaboration},
  author={Tan, Huajie and Hao, Xiaoshuai and Chi, Cheng and Lin, Minglan and Lyu, Yaoxu and Cao, Mingyu and Liang, Dong and Chen, Zhuo and Lyu, Mengsi and Peng, Cheng and others},
  journal={arXiv preprint arXiv:2505.03673},
  year={2025}
}

@article{yang2025efficientnav,
  title={EfficientNav: Towards On-Device Object-Goal Navigation with Navigation Map Caching and Retrieval},
  author={Yang, Zebin and Zheng, Sunjian and Xie, Tong and Xu, Tianshi and Yu, Bo and Wang, Fan and Tang, Jie and Liu, Shaoshan and Li, Meng},
  journal={arXiv preprint arXiv:2510.18546},
  year={2025}
}

@article{yu2025memagent,
  title={MemAgent: Reshaping Long-Context LLM with Multi-Conv RL-based Memory Agent},
  author={Yu, Hongli and Chen, Tinghong and Feng, Jiangtao and Chen, Jiangjie and Dai, Weinan and Yu, Qiying and Zhang, Ya-Qin and Ma, Wei-Ying and Liu, Jingjing and Wang, Mingxuan and others},
  journal={arXiv preprint arXiv:2507.02259},
  year={2025}
}

@inproceedings{yao2025cacheblend,
  title={CacheBlend: Fast large language model serving for RAG with cached knowledge fusion},
  author={Yao, Jiayi and Li, Hanchen and Liu, Yuhan and Ray, Siddhant and Cheng, Yihua and Zhang, Qizheng and Du, Kuntai and Lu, Shan and Jiang, Junchen},
  booktitle={Proceedings of the Twentieth European Conference on Computer Systems},
  pages={94--109},
  year={2025}
}

@inproceedings{wan2025reca,
  title={Reca: Integrated acceleration for real-time and efficient cooperative embodied autonomous agents},
  author={Wan, Zishen and Du, Yuhang and Ibrahim, Mohamed and Qian, Jiayi and Jabbour, Jason and Zhao, Yang and Krishna, Tushar and Raychowdhury, Arijit and Reddi, Vijay Janapa},
  booktitle={Proceedings of the 30th ACM International Conference on Architectural Support for Programming Languages and Operating Systems, Volume 2},
  pages={982--997},
  year={2025}
}

@article{gim2024prompt,
  title={Prompt cache: Modular attention reuse for low-latency inference},
  author={Gim, In and Chen, Guojun and Lee, Seung-seob and Sarda, Nikhil and Khandelwal, Anurag and Zhong, Lin},
  journal={Proceedings of Machine Learning and Systems},
  volume={6},
  pages={325--338},
  year={2024}
}

@article{jin2024ragcache,
  title={Ragcache: Efficient knowledge caching for retrieval-augmented generation},
  author={Jin, Chao and Zhang, Zili and Jiang, Xuanlin and Liu, Fangyue and Liu, Shufan and Liu, Xuanzhe and Jin, Xin},
  journal={ACM Transactions on Computer Systems},
  year={2024},
  publisher={ACM New York, NY}
}

@article{liu2023cachegen,
  title={Cachegen: Fast context loading for language model applications},
  author={Liu, Yuhan and Li, Hanchen and Du, Kuntai and Yao, Jiayi and Cheng, Yihua and Huang, Yuyang and Lu, Shan and Maire, Michael and Hoffmann, Henry and Holtzman, Ari and others},
  journal={CoRR},
  year={2023}
}

@inproceedings{kwon2023efficient,
  title={Efficient memory management for large language model serving with pagedattention},
  author={Kwon, Woosuk and Li, Zhuohan and Zhuang, Siyuan and Sheng, Ying and Zheng, Lianmin and Yu, Cody Hao and Gonzalez, Joseph and Zhang, Hao and Stoica, Ion},
  booktitle={Proceedings of the 29th symposium on operating systems principles},
  pages={611--626},
  year={2023}
}

@article{choi2024lota,
  title={Lota-bench: Benchmarking language-oriented task planners for embodied agents},
  author={Choi, Jae-Woo and Yoon, Youngwoo and Ong, Hyobin and Kim, Jaehong and Jang, Minsu},
  journal={arXiv preprint arXiv:2402.08178},
  year={2024}
}

@inproceedings{shridhar2020alfred,
  title={Alfred: A benchmark for interpreting grounded instructions for everyday tasks},
  author={Shridhar, Mohit and Thomason, Jesse and Gordon, Daniel and Bisk, Yonatan and Han, Winson and Mottaghi, Roozbeh and Zettlemoyer, Luke and Fox, Dieter},
  booktitle={Proceedings of the IEEE/CVF conference on computer vision and pattern recognition},
  pages={10740--10749},
  year={2020}
}

@article{xu2025mem,
  title={A-mem: Agentic memory for llm agents},
  author={Xu, Wujiang and Mei, Kai and Gao, Hang and Tan, Juntao and Liang, Zujie and Zhang, Yongfeng},
  journal={arXiv preprint arXiv:2502.12110},
  year={2025}
}

@article{yao2023retroformer,
  title={Retroformer: Retrospective large language agents with policy gradient optimization},
  author={Yao, Weiran and Heinecke, Shelby and Niebles, Juan Carlos and Liu, Zhiwei and Feng, Yihao and Xue, Le and Murthy, Rithesh and Chen, Zeyuan and Zhang, Jianguo and Arpit, Devansh and others},
  journal={arXiv preprint arXiv:2308.02151},
  year={2023}
}

@article{team2024qwen2,
  title={Qwen2 technical report},
  author={Team, Qwen and others},
  journal={arXiv preprint arXiv:2407.10671},
  volume={2},
  number={3},
  year={2024}
}

@article{zhang2025memgen,
  title={MemGen: Weaving Generative Latent Memory for Self-Evolving Agents},
  author={Zhang, Guibin and Fu, Muxin and Yan, Shuicheng},
  journal={arXiv preprint arXiv:2509.24704},
  year={2025}
}

@article{liu2025freekv,
  title={FreeKV: Boosting KV Cache Retrieval for Efficient LLM Inference},
  author={Liu, Guangda and Li, Chengwei and Ning, Zhenyu and Lin, Jing and Yao, Yiwu and Ke, Danning and Guo, Minyi and Zhao, Jieru},
  journal={arXiv preprint arXiv:2505.13109},
  year={2025}
}

@article{chen2025retroinfer,
  title={RetroInfer: A Vector-Storage Approach for Scalable Long-Context LLM Inference},
  author={Chen, Yaoqi and Zhang, Jinkai and Lu, Baotong and Zhang, Qianxi and Zhang, Chengruidong and Luo, Jingjia and Liu, Di and Jiang, Huiqiang and Chen, Qi and Liu, Jing and others},
  journal={arXiv preprint arXiv:2505.02922},
  year={2025}
}

@article{shi2025look,
  title={Look Back to Reason Forward: Revisitable Memory for Long-Context LLM Agents},
  author={Shi, Yaorui and Chen, Yuxin and Wang, Siyuan and Li, Sihang and Cai, Hengxing and Gu, Qi and Wang, Xiang and Zhang, An},
  journal={arXiv preprint arXiv:2509.23040},
  year={2025}
}

@article{zheng2023efficiently,
  title={Efficiently Programming Large Language Models using SGLang.},
  author={Zheng, Lianmin and Yin, Liangsheng and Xie, Zhiqiang and Huang, Jeff and Sun, Chuyue and Yu, Cody\_Hao and Cao, Shiyi and Kozyrakis, Christos and Stoica, Ion and Gonzalez, Joseph E and others},
  year={2023},
  publisher={arXiv}
}

@article{hu2024epic,
  title={EPIC: Efficient Position-Independent Caching for Serving Large Language Models},
  author={Hu, Junhao and Huang, Wenrui and Wang, Weidong and Wang, Haoyi and Hu, Tiancheng and Zhang, Qin and Feng, Hao and Chen, Xusheng and Shan, Yizhou and Xie, Tao},
  journal={arXiv preprint arXiv:2410.15332},
  year={2024}
}

@article{song2020mpnet,
  title={Mpnet: Masked and permuted pre-training for language understanding},
  author={Song, Kaitao and Tan, Xu and Qin, Tao and Lu, Jianfeng and Liu, Tie-Yan},
  journal={Advances in neural information processing systems},
  volume={33},
  pages={16857--16867},
  year={2020}
}

@article{pan2025kvflow,
  title={KVFlow: Efficient Prefix Caching for Accelerating LLM-Based Multi-Agent Workflows},
  author={Pan, Zaifeng and Patel, Ajjkumar and Hu, Zhengding and Shen, Yipeng and Guan, Yue and Li, Wan-Lu and Qin, Lianhui and Wang, Yida and Ding, Yufei},
  journal={arXiv preprint arXiv:2507.07400},
  year={2025}
}

@article{sun2024shadowkv,
  title={Shadowkv: Kv cache in shadows for high-throughput long-context llm inference},
  author={Sun, Hanshi and Chang, Li-Wen and Bao, Wenlei and Zheng, Size and Zheng, Ningxin and Liu, Xin and Dong, Harry and Chi, Yuejie and Chen, Beidi},
  journal={arXiv preprint arXiv:2410.21465},
  year={2024}
}

@article{cao2025sparse,
  title={Sparse Attention across Multiple-context KV Cache},
  author={Cao, Ziyi and Si, Qingyi and Zhang, Jingbin and Liu, Bingquan},
  journal={arXiv preprint arXiv:2508.11661},
  year={2025}
}

@article{liu2025chunkkv,
  title={Chunkkv: Semantic-preserving kv cache compression for efficient long-context llm inference},
  author={Liu, Xiang and Tang, Zhenheng and Dong, Peijie and Li, Zeyu and Liu, Yue and Li, Bo and Hu, Xuming and Chu, Xiaowen},
  journal={arXiv preprint arXiv:2502.00299},
  year={2025}
}

@article{hu2025efficient,
  title={Efficient Long-Context LLM Inference via KV Cache Clustering},
  author={Hu, Jie and Wang, Shengnan and He, Yutong and Gong, Ping and Yi, Jiawei and Zhang, Juncheng and Bai, Youhui and Chen, Renhai and Zhang, Gong and Li, Cheng and others},
  journal={arXiv preprint arXiv:2506.11418},
  year={2025}
}

@article{sun2025breaking,
  title={Breaking the Boundaries of Long-Context LLM Inference: Adaptive KV Management on a Single Commodity GPU},
  author={Sun, He and Li, Li and Xiao, Mingjun and Xu, Chengzhong},
  journal={arXiv preprint arXiv:2506.20187},
  year={2025}
}

@inproceedings{song2023llm,
  title={Llm-planner: Few-shot grounded planning for embodied agents with large language models},
  author={Song, Chan Hee and Wu, Jiaman and Washington, Clayton and Sadler, Brian M and Chao, Wei-Lun and Su, Yu},
  booktitle={Proceedings of the IEEE/CVF international conference on computer vision},
  pages={2998--3009},
  year={2023}
}

@article{lin2024awq,
  title={Awq: Activation-aware weight quantization for on-device llm compression and acceleration},
  author={Lin, Ji and Tang, Jiaming and Tang, Haotian and Yang, Shang and Chen, Wei-Ming and Wang, Wei-Chen and Xiao, Guangxuan and Dang, Xingyu and Gan, Chuang and Han, Song},
  journal={Proceedings of machine learning and systems},
  volume={6},
  pages={87--100},
  year={2024}
}

@inproceedings{wei2025lightmamba,
  title={Lightmamba: Efficient mamba acceleration on fpga with quantization and hardware co-design},
  author={Wei, Renjie and Xu, Songqiang and Zhong, Linfeng and Yang, Zebin and Guo, Qingyu and Wang, Yuan and Wang, Runsheng and Li, Meng},
  booktitle={2025 Design, Automation \& Test in Europe Conference (DATE)},
  pages={1--7},
  year={2025},
  organization={IEEE}
}

\end{document}